\documentclass[11pt,a4paper]{article}

\usepackage[margin=1in]{geometry}
\usepackage[T1]{fontenc}
\usepackage[utf8]{inputenc}
\usepackage{lmodern}
\usepackage{microtype}
\usepackage{graphicx}
\usepackage{multirow}
\usepackage{makecell}
\usepackage{array}
\usepackage{url}
\usepackage{amsmath}
\usepackage{arydshln}
\usepackage{caption}
\usepackage[hidelinks]{hyperref}

\title{Fusing Visual and Textual Representations\\
via Multi-layer Fusing Transformers\\
for Vietnamese Visual Question Answering}

\author{
Cong Phu Nguyen$^{1,2}$\thanks{ORCID: \href{https://orcid.org/0009-0007-3344-3499}{0009-0007-3344-3499}}
\and
Huy Tien Nguyen$^{1,2}$\thanks{ORCID: \href{https://orcid.org/0000-0002-9948-1048}{0000-0002-9948-1048}}
\and
Tung Le$^{1,2}$\thanks{Corresponding author: lttung@fit.hcmus.edu.vn. ORCID: \href{https://orcid.org/0000-0002-9900-7047}{0000-0002-9900-7047}}
}

\date{}

\begin{document}
\maketitle

\begin{center}
\small
$^1$Faculty of Information Technology, University of Science, Ho Chi Minh City, Vietnam\\
$^2$Vietnam National University, Ho Chi Minh City, Vietnam\\[2pt]
\texttt{21C11018@student.hcmus.edu.vn}\\
\texttt{\{ntienhuy,lttung\}@fit.hcmus.edu.vn}
\end{center}

\begin{abstract}
In recent decades, artificial intelligence has made significant progress in understanding and interacting with images. One of the important applications of this technology is Visual Question Answering (VQA), a research field that requires computers to understand and answer questions about images in a natural manner. Despite extensive research and development in VQA for English, there have been very few similar efforts made for other languages, especially Vietnamese. This gap presents a significant challenge and opportunity for the advancement of VQA technology in the Vietnamese language context. By bridging this gap, the field of Vietnamese VQA not only enriches the diversity of research in artificial intelligence but also enables practical applications in various domains, such as education, healthcare, and entertainment, catering to Vietnamese-speaking populations worldwide. Thus, the exploration and development of Vietnamese VQA systems hold immense potential for advancing both research and practical applications in the intersection of computer vision and natural language processing. In this paper, we propose a Multi-layer Fusing Transformer model utilizing a cross attention module to combine multiple modality features of images and texts from different layers in an aggregated representation. Our architecture allows us extract information from low level to high level. Through detailed experiments and ablation studies, our model achieves promising results against the competitive baselines in ViVQA dataset for Vietnamese language.
\end{abstract}


\section{Introduction}
\label{sec:intro}

In everyday life, humans effortlessly receive and interpret information from multiple sources simultaneously, including visuals, sounds, and textual data. For machines to mirror this complex human capability, they must be adept at processing a variety of signal types concurrently. The field of multi-modal research, especially tasks that combine vision and language, has seen significant growth in recent years, reflecting the increasing interest in understanding and developing systems that can interpret and respond to combined data inputs. Among the myriad of tasks within the realm of image and text multi-modal processing, activities such as Visual Question Answering (VQA), image captioning, text-to-image synthesis, and cross-modal retrieval have emerged as key areas of focus. Notably, Visual Question Answering (VQA) represents a cutting-edge intersection of research and practical application, showcasing the potential of integrating visual and linguistic features. VQA systems are designed to analyze images and comprehend questions about them, synthesizing information from both domains to produce accurate and relevant answers. This ability not only highlights advancements in artificial intelligence (AI) but also opens up new pathways for human-computer interaction, making VQA a critical area of study in the broader field of AI research.



Vietnamese is a widely spoken language with a rich visual culture, making it an important target for developing language-enabled AI systems. However, the development of VQA systems for Vietnamese faces unique challenges due to linguistic nuances, cultural contexts, and the scarcity of annotated datasets in the Vietnamese language. Existing VQA models trained on English-centric datasets may not generalize well to Vietnamese due to language differences and semantic variations. 

Therefore, the primary objective of this research is to design and develop a Vietnamese Visual Question Answering (VQA) system capable of understanding and accurately answering questions about visual content in the Vietnamese language. This system will leverage state-of-the-art techniques in computer vision and natural language processing, adapted and optimized for the Vietnamese language.

This study introduces a novel method designed to enhance the performance of Vietnamese Visual Question Answering (VQA) systems. By addressing the specific challenges of processing visual content and natural language in the Vietnamese language, our proposed method aims to significantly improve the accuracy and reliability of VQA systems tailored for Vietnamese-speaking users. This advancement not only contributes to bridging the gap in language-centric AI technologies but also opens up new possibilities for the development of inclusive and culturally relevant AI applications for Vietnamese communities.

In prior models within the realm of deep learning, the emphasis has predominantly been on employing architectures featuring multiple hidden layers designed to meticulously extract and articulate both visual and linguistic features. Traditionally, these models have harnessed solely the attributes derived from the terminal layer as semantically enriched inputs for ensuing predictive analyses. The efficacy and robustness of such deep, multi-layered constructs are well-established and validated through extensive research across various domains. However, when addressing multi-modal challenges that necessitate the concurrent assimilation of both imagery and linguistic data, a sophisticated approach is not only requisite for the adept representation of features but also imperative for their harmonious integration. This integration aims to engender a more profound semantic synthesis. The conventional strategy of isolating features to the concluding layers of deep learning frameworks, thereby processing visual and textual streams in a segregated manner, inadvertently mitigates the potential for dynamic interaction between images and textual queries throughout the model's learning trajectory.

In light of these considerations, our research introduces the innovative Multi-layer Fusing Transformer (MFT) model. This model is specifically architected to not only represent but also intricately weave together visual and textual features across the entirety of its processing layers. By doing so, the MFT model facilitates a more integrated and continuous interaction between these two pivotal dimensions. This novel approach marks a significant advancement in the field of Vietnamese Visual Question Answering, promising to elevate the efficacy of models tasked with deciphering and responding to complex visual-textual inquiries. Through this cutting-edge methodology, our work sets a new benchmark in multi-modal AI research, particularly in enhancing the comprehension and interpretative capabilities of systems engaging with the rich tapestry of Vietnamese visual and textual content.



Section 2 reviews the background and relevant studies in Visual Question Answering (VQA) for Vietnamese, highlighting challenges and opportunities. Section 3 introduces our proposed VQA model for Vietnamese, detailing its architecture, components, and training methodology. Section 4 describes our experimental setup and evaluation, including the dataset, metrics, and comparative analysis with baseline models. Section 5 concludes with key findings, contributions to VQA research in Vietnamese, and future research directions.
\section{Related Works}
\label{sec:rw}

In recent years, interest in Visual Question Answering (VQA) in the Vietnamese language has significantly increased, with numerous studies exploring its potential and proposing new datasets as well as methods. In 2021, Tran and colleagues \cite{tran2021vivqa} introduced the first ViVQA dataset for VQA in Vietnamese, which has since become the standard in subsequent research.

With the rapid development of technology and computers in recent years, Deep Neural Networks (DNNs) have achieved significant success in various applications. In 2017, the paper "Attention is all you need" by the authors Vaswani et al. \cite{transformer} from Google brought a new wave to the field of natural language processing research. The authors introduced the architecture of Transformers, which, instead of following traditional approaches to text representation, utilizes the self-attention mechanism to estimate the weight of relationships between words in a text and then aggregates the representative information of a word with the weighted information of all words in the text. This implies that in the encoding process, the Transformers architecture can simultaneously process the entire input text without the need for recurrent iterations. At the end of 2018, based on the Transformers architecture, the pre-trained model BERT \cite{BERT} was announced. BERT was proposed with the ability to train using semi-supervised learning methods, making model training easier and more effective in terms of semantic representation of sentences, marking a new milestone in building transferable language models. With the success of the Transformers architecture in natural language processing, many models based on this architecture have been applied to the field of computer vision, such as Vision Transformer (ViT) \cite{VIT} and BeiT \cite{bao2021beit}, yielding promising results compared to models using CNNs.

Current approaches to the Vietnamese VQA problem utilize language and vision models trained on Vietnamese datasets. Notable examples include models like PAT \cite{nguyen2023pat}, BARTPhoBEiT \cite{tran2023bartphobeit} and Multi-vision Contextual Attention (MCA) model \cite{nguyen2022combining}, which are considered state-of-the-art (SOTA) for the VQA task on the ViVQA dataset \cite{tran2021vivqa}. While the PAT model utilizes Parallel Attention with a hierarchical feature extractor for questions, the BARTPhoBeiT model employs multiway transformers on the BardPho \cite{tran2021bartpho} and BeiT \cite{bao2021beit} models. Multi-vision Contextual Attention model simultaneously learns both global and local contextual visual features of the image and uses multi-branch contextual attention fusion that hierarchically integrates visual and textual features.

The findings from \cite{erhan2009visualizing} confirm the intuitions researchers had about the hierarchical representations learned by deep architectures. Specifically, it reveals that the units in higher layers represent features that are significantly more complex and correspond to combinations of features from lower layers. And according to \cite{de2020s}, different layers of BERT exhibit various specializations, suggesting that it could be more beneficial to integrate information from multiple layers rather than selecting a single layer based solely on its overall performance.

\section{Our model}
\label{sec:method}

\begin{figure}[ht!]
\centering
\includegraphics[width=10cm,height=15cm,keepaspectratio]{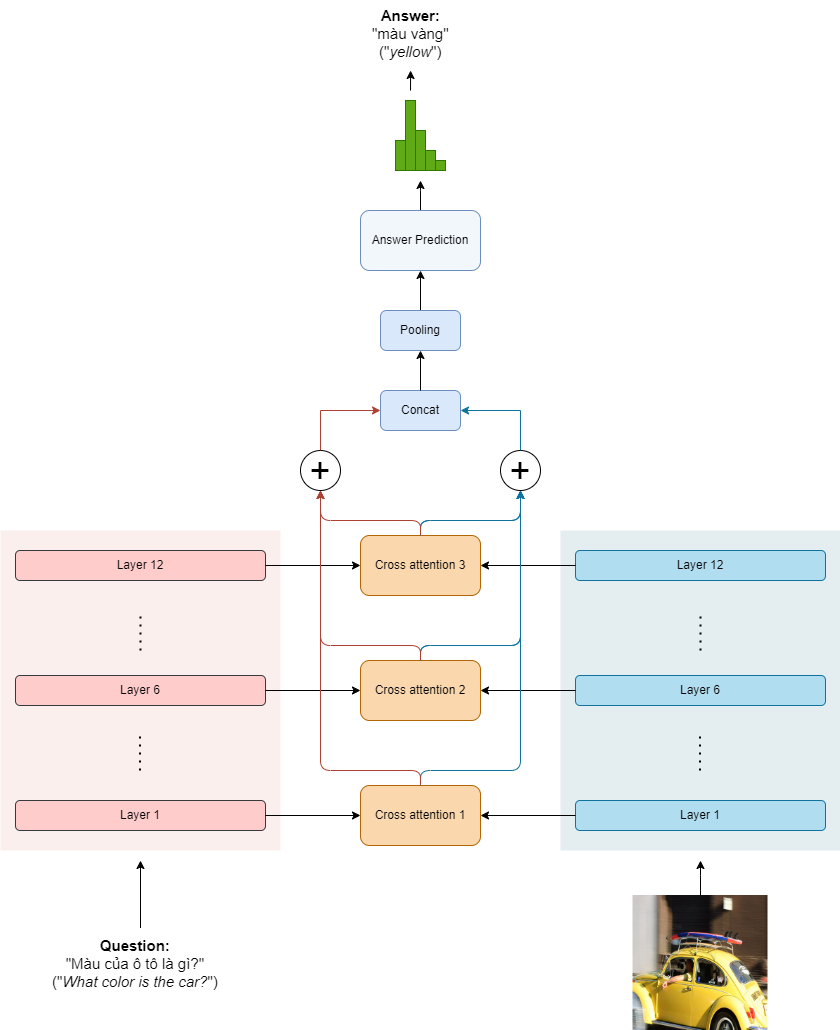}
\caption{Multi-layer Fusing Transformer:  use cross-attention on different layers of pre-trained language model and vision model} \label{figMFT}
\end{figure}

We propose a new method utilizing cross-attention across different layers, called Multi-layer Fusing Transformer (MFT). Additionally, to ensure the model performs well on Vietnamese, we leverage language models that have been pretrained on large Vietnamese datasets. The MFT model consists of 4 main components: the Language Embedding Module, the Visual Embedding Module, the Cross-attention Module, and the Answer Selector (see Fig. \ref{figMFT}). The following section will describe in detail the architecture of the model.

\subsection{Language Embedding Module}

We use PhoBert\cite{phobert}, the pre-trained language model utilized as the Language Embedding Module in our Visual Question Answering (VQA) system. PhoBERT is built upon the Transformer architecture, a powerful neural network architecture renowned for its effectiveness in natural language processing tasks. The Transformer model consists of multiple layers of self-attention mechanisms and feedforward neural networks, allowing it to capture long-range dependencies and contextual information from input sequences. Specifically, PhoBERT employs a multi-layer bidirectional Transformer encoder architecture. Each layer of the encoder consists of self-attention mechanisms, which enable the model to weigh the importance of different words in the input sequence based on their contextual relevance. Additionally, each layer includes feedforward neural networks for feature transformation and activation functions for non-linear mapping. The architecture of PhoBERT allows it to generate contextualized embeddings for words and sentences, capturing the semantic relationships and nuances of the Vietnamese language effectively.

PhoBERT is pre-trained on a large corpus of Vietnamese text data using self-supervised learning techniques. The combination of the Transformer architecture and the self-supervised learning objectives enables PhoBERT to learn rich and meaningful representations of the Vietnamese language, making it well-suited for various downstream tasks, including Visual Question Answering.

\subsection{Visual Embedding Module}
The design of ViT draws inspiration from the success of the BERT\cite{BERT} model in natural language processing. Just as BERT revolutionized NLP by capturing bidirectional context in text sequences, ViT leverages the Transformer architecture to capture global and local context in visual data. Specifically, ViT adapts the self-attention mechanism from BERT to process image patches, enabling it to model complex relationships among visual elements effectively. By treating images as sequences of patches and applying self-attention across these patches, ViT achieves state-of-the-art performance in various computer vision tasks.

\subsection{Cross-attention Module}

\begin{figure}[ht!]
\centering
\includegraphics[width=0.9\textwidth]{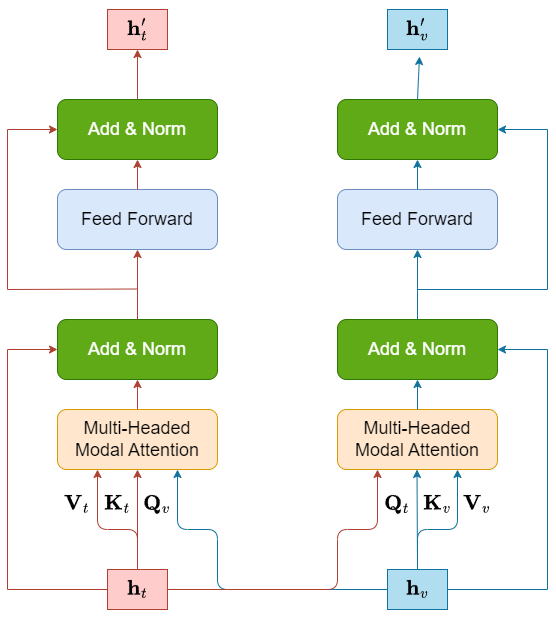}
\caption{Architecture of Cross-Attention} \label{figCA}
\end{figure}

Taking inspiration from the VilBERT\cite{lu2019vilbert}, we employ a cross-attentional transformer layer to effectively integrate information between the question and the image, yielding a representation vector used for answer classification. Each layer has two sub-layers. The initial layer comprises a multi-head cross-attention mechanism, while the subsequent layer consists of a straightforward, fully connected feed-forward network applied position-wise. In the first layer, one modality’s representation is used as the Key and Value vectors, while the other modality’s representation serves as the Query vector. Take the cross-attention of the text branch as an example, the query $\mathbf{Q_v}$ is generated from the image intermediate representations, while the key $\mathbf{K_t}$ and value $\mathbf{V_t}$ is generated from text intermediate representations, the computation can be formulated as:

\begin{equation}
\label{eq:comb}
    Attention(\mathbf{Q_v, K_t, V_t}) = Softmax\left (\frac{\mathbf{Q_v {K_t}^T}}{\sqrt{d_k}}  \right )\mathbf{V_t}, 
\end{equation}

where the subscript t refers to text representations and v for visual representations, respectively. \(\mathbf{K_t}^{T}\) is the transpose of the matrix \(\mathbf{K_t}\). The $d_k$ represents the attention dimension. Using cross-attention fusion layers, both modalities can acquire pertinent and supplementary contextual representations from each other. We utilize a residual connection surrounding each of the two sub-layers, succeeded by layer normalization.

We perform cross-attention taking as input the outputs of language and image processing models at different layers. To integrate information from simple to complex from the two models, representation vectors at the beginning, middle, and end positions (corresponding to layers 0, 6, 12) are fed into the cross-attention layers. The representation vectors are averaged for both the text and vision parts, and the vector corresponding to the position of the [CLS] token of each part is concatenated together and passed through a pooling layer followed by a softmax operation to obtain probabilities for each answer.

\subsection{Answer Selector}
The Answer Selector module fuses visual features \(h'_v\) and linguistic features \(h'_t\) to produce fused features \(h'_f\). These fused features are then projected into the vocabulary space, resulting in a probabilistic vector that indicates the most likely candidate answer.
\section{Experiments}
In this section, we conduct a series of experiments to evaluate the performance of our proposed MFT models alongside established state-of-the-art (SOTA) models. Our experiments are conducted on the ViVQA dataset \cite{tran2021vivqa}, chosen for its relevance and suitability for assessing the efficacy of visual question answering systems. We provide comprehensive details regarding the implementation process and the selection of evaluation metrics, ensuring a fair and rigorous comparison between our models and existing benchmarks on the same dataset.
\label{sec:exp}

\subsection{Dataset}
The ViVQA dataset, created for Vietnamese Visual Question Answering (VQA), addresses the lack of resources in this domain despite the growing interest in VQA within natural language processing and computer vision communities. It consists of 10,328 images paired with 15,000 questions and answers in Vietnamese, facilitating the evaluation of VQA models tailored for the Vietnamese language. The dataset is divided randomly into training and test sets at an 8:2 ratio. These questions are categorized into types such as Object, Number, Color, and Location, and distributed across the train and test sets as shown in the Table~\ref{tab:dataset}

\begin{table}
\caption{Statistics of the ViVQA dataset}
\label{tab:dataset}
\begin{center}
\begin{tabular}{l|l|l}
\Xhline{2\arrayrulewidth}
     &  {\bfseries Training} & {\bfseries Test}\\
\Xhline{2\arrayrulewidth}
    {\bfseries Number of samples}       & 11,999   & 3001\\
\Xhline{2\arrayrulewidth}
    {\bfseries Object}                  & 41.55\%   & 41.56\%\\
    {\bfseries Number}                  & 14.81\%   & 14.81\%\\
    {\bfseries Color}                   & 20.82\%   & 20.82\%\\
    {\bfseries Location}                & 22.82\%   & 22.82\%\\
\Xhline{2\arrayrulewidth}

\end{tabular}
\end{center}
\end{table}

\subsection{Evaluation Metrics}
Following  Rajpurkar et al.\cite{rajpurkar2016squad}, we use two evaluation metrics: F1 score and Exact Match. The F1 score serves as a crucial metric in assessing the performance of visual question answering models. It quantifies the average overlap between predicted and ground truth answers, treating them as bags of tokens to compute their F1 score. Specifically, the F1-score for each answer is computed based on the tokens of both the gold answer (GA) and the predicted answer (PA), ensuring a comprehensive evaluation. This calculation is performed for each question within the dataset, with the overall F1 score being averaged across all questions to provide a comprehensive assessment of model performance. In the context of Vietnamese VQA, F1 scores are calculated based on tokenized versions of the answers, facilitating precise evaluation and comparison across different models and datasets. Exact Match evaluates the proportion of predictions that exactly match any of the ground truth answers.

\begin{equation}
\label{eq:precision}
Precision\left ( P \right ) = \frac{GA\cap PA}{PA}
\end{equation}

\begin{equation}
\label{eq:recall}
Recall\left ( P \right ) = \frac{GA\cap PA}{GA}
\end{equation}

\begin{equation}
\label{eq:f1}
F1 = \frac{2 \times P \times R}{P + R}
\end{equation}

\subsection{Experimental Settings}
In our model, we take advantage of pre-trained models for extracting features from images and questions. The name of initial version in our feature extraction modules are presented in Table~\ref{tab:setting}. Besides, we also provide more details to make the reproducing process easier. For Multi-head Attention and Transformer components, we also show the range of our tuning process. The experiments were performed on a computing system with a CPU of Intel(R) Xeon(R) E5-2696 v3, a GPU of RTX3060, and a RAM of 48GB.

\begin{table}
\caption{Detail of Component Setting}
\label{tab:setting}
\begin{center}
\begin{tabular}{l|l}
\Xhline{2\arrayrulewidth}
    {\bfseries Component} &  {\bfseries Value}\\
\Xhline{2\arrayrulewidth}
    {\bfseries Visual Embedding Module}     & google/vit-base-patch16-384\\
    {\bfseries Lanuguage Embedding Module}  & vinai/phobert-base-v2\\
    {\bfseries No. Connections}             & i $\in$ \{\(0, 0\), \(6, 6\), \(12, 12\)\}\\
    {\bfseries Batch size}                  & 32\\
    {\bfseries Optimizer}                   & AdamW(lr = 1e-4, eps = 1e-8)\\
    {\bfseries Learning rate Scheduler}     & OneCycleLR(lr = 1e-4, pct\_start = 0.05)\\
\Xhline{2\arrayrulewidth}

\end{tabular}
\end{center}
\end{table}

\subsection{Results}

Because there are few studies on Visual Question Answering (VQA) for Vietnamese, there are also few methods for this problem since the ViVQA dataset was published. In this section, we will compare the MFT model with recent SOTA models on the ViVQA dataset, including the Parallel Attention Transformer (PAT) \cite{nguyen2023pat}, BARTPhoBeiT \cite{tran2023bartphobeit} and Multi-vision Contextual Attention (MCA) model \cite{nguyen2022combining}. For these models, the results are extracted from the publication of these models. The details of the comparison are presented in Table~\ref{tab:result}.

\begin{table}
\caption{The comparison of our model and competitive baselines (test-standard)}
\label{tab:result}
\begin{center}

\begin{tabular}{m{5cm} >{\centering\arraybackslash}m{1.5cm} >{\centering\arraybackslash}m{1.5cm} >{\centering\arraybackslash}m{1.5cm} >{\centering\arraybackslash}m{1.5cm}}
\Xhline{2\arrayrulewidth}
    \textbf{Model} & \textbf{Precision} & \textbf{Recall} & \textbf{F1-score} & \textbf{Exact Match}\\
\Xhline{2\arrayrulewidth}
    \( {ViHieCoAtt - syllable}_{100 dims}\) \cite{tran2023bartphobeit}      & 0.2271 & 0.2954 & 0.2234 & -\\
    \( {ViHieCoAtt - word}_{100 dims}\) \cite{tran2023bartphobeit}          & 0.2049 & 0.3044 & 0.2256 & -\\
    \( {ViHieCoAtt - syllable}_{300 dims}\) \cite{tran2023bartphobeit}      & 0.2185 & 0.3104 & 0.2277 & -\\
    \( {ViHieCoAtt - word}_{300 dims}\) \cite{tran2023bartphobeit}          & 0.2192 & 0.3236 & 0.2436 & -\\
\Xhline{2\arrayrulewidth}
    \( {PAT}\) \cite{nguyen2023pat}                                         & -      & -      & -      & 0.6055\\
    \( {MCA}\) \cite{nguyen2022combining}     & -      & -      & -      & 0.6076\\
    \( {BARTPhoBeit - base}_{syllable}\) \cite{tran2023bartphobeit}         & 0.6684 & 0.6718 & 0.6610 & -\\
    \( {BARTPhoBeit - base}_{word}\) \cite{tran2023bartphobeit}             & 0.6593 & 0.6646 & 0.6518 & -\\
    \( {BARTPhoBeit - large}_{syllable}\) \cite{tran2023bartphobeit}        & 0.6931 & 0.6858 & 0.6777 & -\\
    \( {BARTPhoBeit - large}_{word}\) \cite{tran2023bartphobeit}            & 0.6755 & 0.6749 & 0.6643 & -\\
\Xhline{2\arrayrulewidth}
    \(MFT (our)\)      & \textbf{0.7057} & \textbf{0.7060} & \textbf{0.7044} & \textbf{0.6485}\\
\end{tabular}
\end{center}
\end{table}

In comparison to existing models in the field, our proposed Multi-layer Fusing Transformer (MFT) model exhibits notable advantages across various performance metrics. While current state-of-the-art (SOTA) models have demonstrated commendable performance in Vietnamese Visual Question Answering (VQA) tasks, our MFT model surpasses them in terms of both F1-score and Exact metric. Specifically, when compared to the best-performing version of the BARTPhoBeit model, MFT showcases a significant improvement, achieving a 4\% higher F1-score. Moreover, in terms of the Exact metric, our proposed model exhibits a similarly substantial lead of 4\% over the PAT model. These results underscore the effectiveness and superiority of the MFT model in capturing intricate semantic relationships between visual and textual inputs, thereby enabling more accurate and nuanced responses in VQA tasks. Additionally, the innovative cross-attention mechanism integrated across different layers within the MFT architecture enhances its capability to effectively fuse information from both modalities, leading to superior performance outcomes. Overall, the comparative analysis highlights the MFT model's potential as a leading solution in advancing the state-of-the-art in Vietnamese VQA research.

\subsection{Ablation Studies}
In this section, we experiment with the ability to combine cross-attention modules at different layer positions as well as the number of cross-attention layers on the language model and image processing model.

\begin{table}[ht!]
\caption{The effect of using cross-attention on different layer pairs}
\label{tab:abl1}
\begin{center}

\begin{tabular}{m{5cm} >{\centering\arraybackslash}m{1.5cm} >{\centering\arraybackslash}m{1.5cm} >{\centering\arraybackslash}m{1.5cm} >{\centering\arraybackslash}m{1.5cm}}
\Xhline{2\arrayrulewidth}
    \textbf{Connection layers} & \textbf{Precision} & \textbf{Recall} & \textbf{F1-score} & \textbf{Exact Match}\\
\Xhline{2\arrayrulewidth}
    (0, 0), (6, 6), (12, 12)         & \textbf{0.7057} & \textbf{0.7060} & \textbf{0.7044} & \textbf{0.6485}\\
    (0, 12), (6, 6), (12, 0)              & 0.7007 & 0.7003 & 0.6990 & 0.6415\\
    (0, 6), (6, 12), (12, 0)              & 0.6956 & 0.6954 & 0.6939 & 0.6351\\
    (0, 12), (6, 0), (12, 6)              & 0.7012 & 0.7004 & 0.6992 & 0.6415\\
    (0, 0), (1, 1), (2, 2)                & 0.6217 & 0.6207 & 0.6192 & 0.5548\\
    (5, 5), (6, 6), (7, 7)                & 0.6602 & 0.6587 & 0.6578 & 0.5975\\
    (10, 10), (11, 11), (12, 12)          & 0.6808 & 0.6820 & 0.6798 & 0.6198\\
\Xhline{2\arrayrulewidth}
\end{tabular}
\end{center}
\end{table}

The findings indicate that integrating layers at non-matching positions does not yield favorable outcomes. Furthermore, exclusively combining layers either at the beginning, the middle, or at the end does not produce satisfactory results compared to distributing positions evenly at the beginning, middle, and end of the models.

\begin{table}[ht!]
\caption{The effect of using cross-attention on number of cross-attention layers}
\label{tab:abl2}
\begin{center}

\begin{tabular}{m{1cm} >{\centering\arraybackslash}m{3cm} >{\centering\arraybackslash}m{1.5cm} >{\centering\arraybackslash}m{1.5cm} >{\centering\arraybackslash}m{1.5cm} >{\centering\arraybackslash}m{1.5cm}}
\Xhline{2\arrayrulewidth}
    \textbf{Num layers} & \textbf{Connection layers} & \textbf{Precision} & \textbf{Recall} & \textbf{F1-score} & \textbf{Exact Match}\\
\Xhline{2\arrayrulewidth}
    2 & (0, 0), (12, 12)              & 0.7033 & 0.7032 & 0.7017 & 0.6441\\
    3 & (0, 0), (6, 6), (12, 12)     & \textbf{0.7057} & \textbf{0.7060} & \textbf{0.7044} & \textbf{0.6485}\\
    4 & (0, 0), (4, 4), (8, 8), (12, 12)              & 0.7009 & 0.7012 & 0.6996 & 0.6451\\
    5 & (0, 0), (3, 3), (6, 6), (9, 9), (12, 12)      & 0.7054 & 0.7053 & 0.7039 & 0.6468\\
\Xhline{2\arrayrulewidth}
\end{tabular}
\end{center}
\end{table}

Conducting experiments with varying numbers of layers follows a similar process as when there are 3 layers. The results in the table represent the best outcomes corresponding to different numbers of layers. From the results table, we can conclude that the model performs best when the number of layers is 3.
\section{Conclusion}
\label{sec:conclusion}

In conclusion, this study presents a novel approach aimed at enhancing the performance of Vietnamese Visual Question Answering (VQA) systems. By addressing the unique challenges associated with processing visual content and natural language in the Vietnamese language, our proposed method significantly improves the accuracy and reliability of VQA systems tailored for Vietnamese-speaking users. The Multi-layer Fusing Transformer (MFT) model, introduced in this study, outperforms existing models in the field across various performance metrics. Specifically, compared to state-of-the-art models such as BARTPhoBeit and PAT, our MFT model demonstrates a notable improvement, achieving a 4\% higher F1-score and Exact metric. These results highlight the effectiveness of the MFT model in capturing complex semantic relationships between visual and textual inputs, leading to more accurate and nuanced responses in VQA tasks. The innovative cross-attention mechanism integrated across different layers within the MFT architecture further enhances its capability to fuse information from both modalities, resulting in superior performance outcomes. Overall, our findings underscore the potential of the MFT model as a leading solution in advancing Vietnamese VQA research, while also paving the way for the development of inclusive and culturally relevant AI applications for Vietnamese communities.

\section*{Acknowledgment}
This research is funded by University of Science, VNU-HCM under grant number
CNTT 2023–07

\end{document}